# Embedded Design of Automatic Pesticide Spraying Robot Control System

AHAMED MUSTAK*1, HONGBIN MA*2, LEPENG SONG*3, and YING JIN*4

*1Beijing Institute of Technology, 5 South Zhong Guan Cun Street, Haidian District, 100081, Beijing, China.
E-mail: mustak@bit.edu.cn, mustak@qq.com
*2Beijing Institute of Technology, 5 South Zhong Guan Cun Street, Haidian District, 100081, Beijing, China.
E-mail: mathmhb@139.com
*3Chongqing University of Science and Technology, 20 East Road, Huxi Avenue, University Town, Shapingba District, 401331, Chongqing, China.
*4Beijing Institute of Technology, 5 South Zhong Guan Cun Street, Haidian District, 100081, Beijing, China.

**Abstract.** In agriculture, crops need to apply pesticide spraying flow control precisely to reduce costs, protect the environment, and increase yield production. Although there have several variable control methods for spraying flow control because indirect control flow techniques and having a slow response could cause inaccuracy and mismanagement, also noted that those systems also suffer from complicated design and debugging, etc. In this paper, an embedded design of the fuzzy PID variable spraying control method is adopted. The experimental results show that the overshoot of Proportional–Integral–Derivative (PID) control is 10.76%, and the overshoot of fuzzy PID control is 7.17% which can meet the requirements of an advanced spray control flow system. For further investigation, a novel spray flow control method based on Programmable Logic Control (PLC) is proposed in this paper.

**Keywords:** Fuzzy PID, MATLAB, PLC，Control System Simulation.

## 1. INTRODUCTION

As farms grow in size, together with the size of the equipment used on them, there is a need for ways to automate processes, before performed by the farmer himself, such as controlling the fields for pests. These tasks are suited for autonomous robots, as they often need many repetitions over a long period and a large area [1]. In most cases, a small agricultural robot would be ineffective in performing farming jobs, as these often need a lot of materials, either to put into the ground, such as seeds or fertilizers or to take from the field during harvest. But when dealing with monitoring and mapping of fields or precision spraying of pesticides, a smaller robot is ideal, as it is gentler on the crops but also to the ground [2]. Often, there is a need for fungicides and pesticides for the optimal growth of the plant and the full life of a plant. Automating tasks within the farm will enable the avoidance of hazardous human exposure to pesticides and can increase the efficiency and productivity of the farm. For the achievement of the desired conditions, the use of fungicides, and pesticides are often done by farmers [3]. At present, spraying a variety of chemical pesticides is the main prevention and control measure to remove pests and diseases in agriculture, and the unscientific use of chemical pesticides will have a serious impact on the atmosphere, soil, crops themselves, and other environmental organisms, and destroy the ecological balance. Furthermore, most of the tasks can be done by autonomous agricultural robots instead of hiring the workers to do miscellaneous work thus it will save the expenses on the labor [13].

Therefore, the common problem with an agricultural autonomous robot is the navigation problem with the decision-making capability used to able the robot fully operated. To facilitate this issue, there are some research has been done to navigate through all the fields [15-18]. As artificial intelligence (AI) starts to emerge, the current robot should be able to navigate the next movement by the adaptation to the surrounding environment and decide which path it will take. The typical method used in the detection is based on the targeted object orientation or repelled signal emits from the sensor itself and then calculates the distance in between it [19-24]. Some certain researchers may focus on UAV-based pesticide spraying, localization and motion control of agriculture robots, pest image identification, and so on [14]. In this paper, the flow loop and pressure loop were designed to improve the pesticide utilization rate. The flow loop was controlled by variable domain fuzzy PID, and the pressure loop was controlled by the fuzzy controller.

Farmers are currently spraying pesticides around their fields. For the effective use of pesticides and precision spraying of pesticides, the farmers are still struggling with those issues. Overcoming these issues and contributing to this sector is the main objective of this project [5-8]. To improve the performance of automatic robots when insecticide spraying, the PLC control system and the adaptive Fuzzy PID control algorithm are combined to study the automatic pesticide spraying robot.

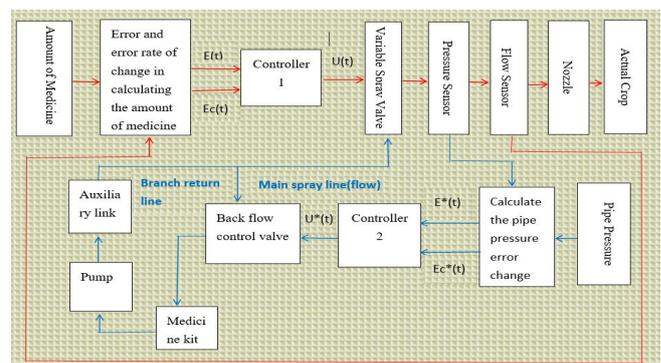

Fig. 1. Structure diagram of precision spraying system

This project presents a technological solution to develop and design the control system with possible high accurate results for pesticide spraying robots. In Figure 1, the outer closed-loop is the flow control (red loop), and the inner closed-loop is the pressure control (blue loop).

## 2. Motivation & Related Work

### 2.1. Motivation

Agriculture is quickly becoming an exciting high-tech industry, drawing new professionals, new companies, and new investors. The technology is developing rapidly, not only advancing the production capabilities of farmers but also advancing robotics and automation technology as we know it. At the heart of this phenomenon is the need for significantly increased production yields [4]. The UN estimates the world population will rise from 7.3 billion today to 9.7 billion in 2050. The world will need a lot more food, and farmers will face serious pressure to keep up with demand.

Agricultural systems globally are now highly reliant on the large-scale application of synthetic pesticides to control weeds, insects, and diseases. An improved understanding of chemical safety and environmental impacts has led to multiple product withdrawals, reducing the number of active ingredients available to farmers. Spraying pesticides and weed killers onto fields is not only wasteful, but it can also severely harm the environment. Robots provide a much more efficient method. Furthermore, ever higher regulation and registration costs have reduced the number of new pesticides entering the agricultural market. There is therefore a global need to find new ways to produce crops that do not require or reduce the use of pesticides. There are now several crop weeding robots that reduce the need for herbicides by deploying camera-guided hoes [9], precision sprayers [10], or lasers [11] to manage weeds. Although in its infancy, this technology shows great promise. In addition, novel sensors deployed on robots can reduce pesticide use by both detecting pests and diseases and precisely targeting the application of insecticides and fungicides. Robots could also be deployed as part of integrated pest management systems, for example, for the accurate and low-cost dispersal of bio-pesticides to counteract crop pests and diseases. The concept of micro-spraying could significantly reduce the amount of herbicide used in crop growing. Micro-spraying robots use computer vision technology to detect weeds and then spray a targeted drop of herbicide onto them. AG BOT II is a solar-powered robot that uses this technique.

### 2.2. Related Work

Agricultural robots are increasing production yields for farmers in various ways. From drones to autonomous tractors to robotic arms, the technology is being deployed in creative and innovative applications. Robots have many fields of application and research development in agriculture. Some examples and prototypes of robots include the Merlin Robot Milker, Rosphere, Harvest Automation, Orange Harvester, lettuce bot, and weeder. One case of large-scale use of robots in farming is the milk bot. It is widespread among British dairy farms because of its efficiency and non-requirement to move. According to David Gardner (chief executive of the Royal Agricultural Society of England), a robot can complete a complicated task if it's repetitive and the robot is allowed to sit in a single place. Furthermore, robots that work on repetitive tasks (e.g. milking) fulfill their role to a consistent and particular standard.

Dr. M.G. Sumithra and G.R. Gayathiri proposed in their paper ―Leaf Disease Diagnosis and Pesticide Spraying Using Agricultural Robot (AGROBOT)‖ that Plant diseases have created an immense post-effect scenario as they can significantly reduce agricultural products in terms of both quality and quantity. Philip J. Sammons, Furukawa Tomonari, and Bulgin Andrew proposed in their paper ―Autonomous Pesticide Spraying Robot for use in a Greenhouse [3]. That is an engineering solution that includes spraying potentially toxic chemicals in the confined space of a hot and steamy glasshouse to the current human health hazards.

China's research in automation in the agricultural sector is faster than ever though China's robot development started late, with the invention of the first industrial robot prototype in 1980. In 1982, Shenyang Institute of Automation developed the first industrial robot in China. But our country is advancing by leaps and bounds in the development of robotics, especially in agriculture. In 2007, to study the navigation path identification method of pesticide spraying robot in the cotton field, Professor Sun Yuanyi took cotton field images collected in the natural environment as the research background and processed them in Lab color space to identify cotton plants from the soil background. In 2010, Professor Cao Zhengyong led the team to design a 3-DOF spray robot control system according to the planting mode and growth characteristics of hedgerow-type plants in the modern greenhouse to improve the effective utilization rate of pesticide application and reduce pesticide residue and chemical pollution. In 2012, Professor Geng Changxing conducted an experimental evaluation on the performance of target spray in an intelligent spraying robot greenhouse. The spray robot uses vision sensors to collect plant and disease images, provide focus location and disease information, and determine the direction of application in the degree of freedom Cartesian coordinate system. In 2017, Professor Diao Zhihua et al. proposed a disease recognition and application algorithm suitable for the application robot system to solve the problems of the low pesticide utilization rate of traditional pesticide application machinery and the failure to detect crop diseases in the application process.

## 3. Methodology

The design and simulation of the controller of the variable spraying system are mainly based on the design of a fuzzy PID controller for the variable spraying control system. The variable theory domain fuzzy PID controller combined with the mathematical model of the variable spray control system to establish a new system by using the MATLAB simulation. Then the PLC software system is constructed according to the controller.

### 3.1. Design of FUZZY PID Controller

Fuzzy PID controller has combined fuzzy control and PID control strategy composed of the controller, combining with the characteristics of variable spray pesticide control system

According to the design and modeling of the variable spray control system, it can be determined that the fuzzy PID controller is mainly for the control of pressure and flow in the spray pipeline. In this section, the design steps of the fuzzy PID controller are determined based on the characteristics of the variable spraying control system, and the pressure and flow in the spray pipeline of the variable spraying control system are controlled and simulated. A PID controller is a control loop mechanism employing feedback that is widely used in industrial control systems and a variety of other applications requiring continuously modulated control. A PID controller continuously calculates an error value e(t) as the difference between the desired set-point (SP) and a measured process variable (PV) and applies a correction based on proportional, integral, and derivative terms. An example is the cruise control on a car, where ascending a hill would lower speed if only constant engine power were applied. The controller's PID algorithm restores the measured speed to the desired speed with minimal delay and overshoot by increasing the power output of the engine.

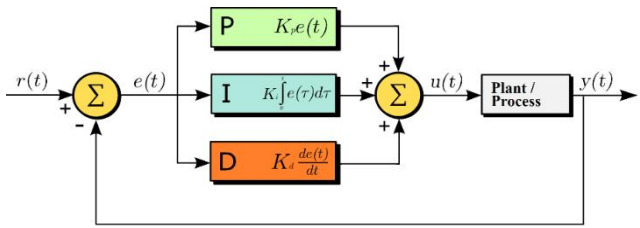

Fig. 2. Block diagram of PID controller

The block diagram in the upper shows the principles of how these terms are generated and applied. It shows a PID controller, which continuously calculates an error value e(t) as the difference between a desired setpoint SP=r(t) and a measured process variable PV=y(t): e(t) = r(t)-y(t), and applies a correction based on proportional, integral, and derivative terms. The controller attempts to minimize the error over time by adjustment of a control variable u(t), such as the opening of a control valve, to a new value determined by a weighted sum of the control terms.

$$u(t) = K_p e(t) + K_i \int_0^t e(t)\,dt + K_d \frac{de(t)}{dt}, \quad (1)$$

In equation (1) where the $K_p$, $K_i$, and $K_d$, all non-negative, denote the coefficients for the proportional, integral, and derivative terms respectively. In the standard form of the equation, $K_i$ and $K_d$ are respectively replaced by $K_p/T_i$ and $K_p T_d$; the advantage of this is that $T_i$ and $T_d$ have some understandable physical meaning, as they represent the integration time and the derivative time respectively.

$$u(t) = K_p \left( e(t) + \frac{1}{T_i} \int_0^t e(t)\,dt + T_d \frac{de(t)}{dt} \right) \quad (2)$$

The PID control scheme is named after its three correcting terms, whose sum constitutes the manipulated variable (MV). The proportional, integral, and derivative terms are summed to calculate the output of the PID controller. Defining u(t) as the controller output, the final form of the PID algorithm is:

$$u(t) = MV(t) = K_p e(t) + K_i \int_0^t e(\tau)\,d\tau + K_d \frac{de(t)}{dt} \quad (3)$$

Where,
Kp is the proportional gain, a tuning parameter, Ki is the integral gain, a tuning parameter, Kd is the derivative gain, a tuning parameter,
e(t)=SP -PV(t) is the error (SP is the setpoint, and PV(t) is the process variable),
t is the time or instantaneous time (the present),
τ (tau) is the variable of integration (takes on values from time 0 to the present t).
Equivalently, the transfer function in the Laplace domain of the PID controller is

$$L(s) = K_p + K_i/s + K_d s, \quad (4)$$

where s is the complex frequency in equation (4).

Based on the foregoing theory, the fuzzy PID controller for flow control of the variable spraying system is designed. The design steps of the controller are as follows:

1. Determine the flow input variation range and output variation range of variable spray spraying system, and then determine the quantization level, quantization factor, and scale factor of flow error and error change rate, and △Kp, △Ki, △Kd.

2. Define fuzzy subsets in the quantization domain of each variable. Firstly, the number of fuzzy subsets is determined, and the language variables of the fuzzy subset are determined. Then, the membership function is selected for each language variable.

3. Determine fuzzy control rules. The principle to be followed in determining fuzzy control rules is to ensure that the output of the controller can make the corresponding dynamic and static performance indexes of the system reach the optimum.

4. Find the fuzzy control table. The fuzzy rule sentence form adopted by the fuzzy controller is "if E is α and EC is β, then U is γ.", where α, β, and γ represent the fuzzy set corresponding to each parameter. The linguistic variable E of the deviation e of the given fertilization amount K and the output fertilization amount L has 7 fuzzy subsets, and the linguistic variable EC of the rate of change of the deviation e has 7 fuzzy subsets. The measured data is obtained by the control expert after experience and cognitive processing, 49 control rules are obtained. The rule table is shown in Table 1.

5. After the second and third steps, the sampling deviation and deviation change rate are substituted into the table of fuzzy control rules to obtain new parameters. After the calculation of the algorithm, the final output quantity is obtained, that is, the control quantity of the system.

6. Establish the flow fuzzy PID control simulation system, analyze the dynamic performance of the flow control of the variable spray spraying system, and then set the quantization factor and scale factor online in combination with the dynamic performance, until the system is adjusted to achieve the ideal control effect.

Table. 1 $K_p$, $K_i$, $K_d$ Fuzzy Control Rules

| $K_p, K_i, K_d$ \ E<br>EC | NB | NM | NS | 0 | PS | PM | PB |
|---|---|---|---|---|---|---|---|
| NB | PB,NB,PS | PB,NB,PS | PM,NB,0 | PM,NM,0 | PS,NM,0 | PS,0,PB | 0, 0, PB |
| NM | PB,NB,NS | PB,NB,NS | PM,NM,NS | PM,NM,NS | PS,NS,0 | 0, 0, NS | 0, 0, PM |
| NS | PM,NM,NB | PM,NM,B | PM,NS,NM | PS,NS,NS | 0, 0, 0 | NS,PS,PS | NM,PS,PM |
| 0 | PM,NM,NB | PS,NS,NM | PS,NS,NM | 0, 0, NS | NS,PS,0 | NM,PS,PS | NM,PM,PM |
| PS | PS,NS,NB | PS,NS,NM | 0, 0, NS | NS,PS,NS | NS,PS,0 | NM,PM,PS | NM,PM,PS |
| PM | 0, 0, NM | PS,NS,NM | PS,PS,NS | NM,PM,NS | NM,PM,0 | NM,PB,PS | NB,PB,PS |
| PB | 0, 0, PS | NS, 0, 0 | NS,PS,0 | NM,PM,0 | NM,PB,0 | NB,PB,PB | NB,PB,PB |

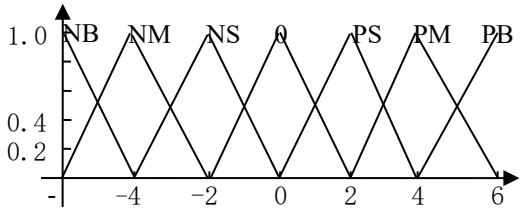

Fig. 3. Membership functions of E, EC and U

The fuzzy controller of this research selects the deviation e of the given fertilizer application amount K and the output fertilizer amount L of the system and the deviation change rate ec as input [12]. According to the simulation of the fertilization system, it is calculated that e∈[0, 0.9], ec∈[-4.67, 0]. The linguistic variables E and EC are obtained after fuzzification of e and ec. Their fuzzy sets are {NB, NM, NS, 0, PS, PM, PB}, and their domains are {-6,-5,- 4,-3,-2,-1,0,1,2,3,4,5,6}, the membership functions of both select triangular membership functions, as shown in Figure 3.

Take the control quantity u as the output, where u∈[-1.127, 0.853]. The fuzzy set of the linguistic variable U of the control quantity is {NB, NM, NS, 0, PS, PM, PB}, and the domain is {-6,-5,-4,-3,-2,-1,0, 1,2,3,4,5,6}, its membership function also chooses a triangular membership function, as shown in Figure 3.

The quantitative factors of the deviation e of the given fertilizer amount K and the output fertilizer amount L and the deviation change rate ec of the variable application granular fertilizer control system are Ke and Kec, respectively, and the proportional factor of the control amount u is Ku. After the linear mapping between the real value range and the linguistic variable universe, here we take Ke=5, Kec=0.8, and Ku=0.45.

According to the design steps of the fuzzy PID controller, the fuzzy PID controller of the variable spraying control system is designed. The structure of fuzzy PID controller of variable spraying control system is shown in Figure 4.

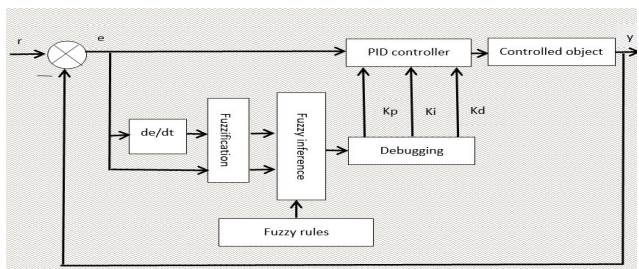

Fig. 4. Structure Diagram of Fuzzy PID Controller

According to the Fuzzy PID control principle, theory, control rules table, and design steps, a conventional fuzzy PID is designed in figure 5;

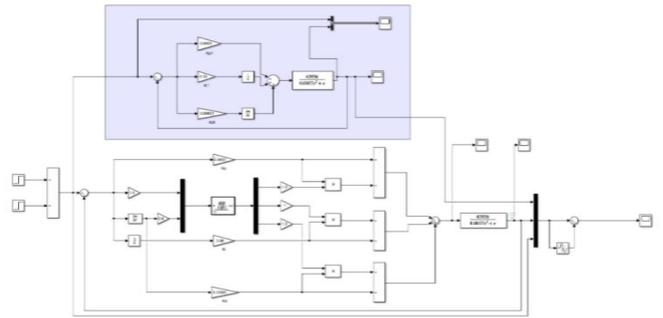

Fig. 5. Fuzzy PID control and conventional PID control simulation structure diagram

### 3.2. PLC Software Architecture Design

The upper computer system of this project is carried out in TIA Portal V15 programming software through SIMATIC WINCC PROFESSIONAL V15. It is more convenient to use WinCC upper computer with PLC S7-1200 software. At the same time, it has many features, such as multi-function, flexibility and simplicity, and strong extensibility.

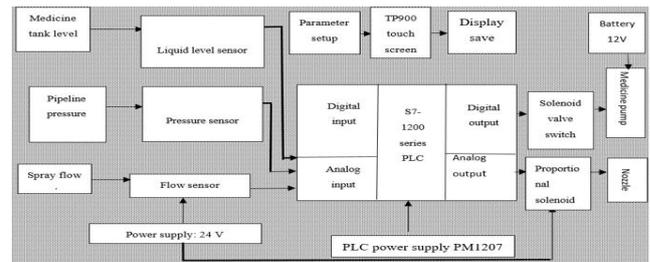

Fig. 6. Diagram of PLC Control System for Variable Spray

The design for the PLC architecture of PID compact is shown if figure 7;

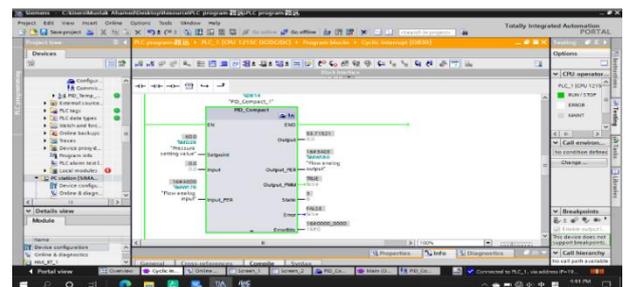

Fig. 7. PLC Architecture Design

Embedded Design of Automatic Pesticide Spraying Robot Control System

## 4. IMPLEMENTATION AND RESULTS

Set up the simulation platform of the variable spray application system, analyze the control performance of fuzzy PID of the variable spray application system according to the simulation results and test results, and then adjust each quantization factor and scale factor of the variable spray application system to achieve the ideal control effect.

With the input and output response of the fuzzy PID controller in Figure 8, it can be seen that, for such a third-order system as the model, the overshoot of the fuzzy PID controller is almost 10% and the adjustment time is the 60s.

$$G(s) = 43956 / 0.0037s^2 + s \qquad (5)$$

Determine the variation range of each input and output variable of the variable spray spraying control system according to the actual needs, and then determine the quantization level, quantization factor and proportion factor of each variable of the variable spray spraying control system and the quantitative rating would take $\{-6, 5, 4, 3, 2, 1, 0, +1, +2, +3, 4, 5, 6\}$. $K_p\ K_i\ K_d$

Define fuzzy subsets in the quantization domain of each variable. the membership function of fuzzy variables such as equation (6) ~ (9).

$$f(x) = \begin{cases} -0.5x - 2 & (x < -5) \\ 0.5x + 3 & (-5 \leq x < -4) \\ -0.5x - 1 & (-4 \leq x < -3) \end{cases} \qquad (6)$$

$$f(x) = \begin{cases} 0.5x + 2 & (-3 \leq x < -2) \\ -0.5x & (-2 \leq x < -1) \\ 0.5x + 1 & (-1 \leq x < 0) \end{cases} \qquad (7)$$

$$f(x) = \begin{cases} -0.5x + 1 & (0 \leq x < 1) \\ 0.5x & (1 \leq x < 2) \\ -0.5x + 2 & (2 \leq x < 3) \end{cases} \qquad (8)$$

$$f(x) = \begin{cases} 0.5x + 1 & (3 \leq x < 4) \\ -0.5x + 3 & (1 \leq x < 2) \\ 0.5x + 2 & (2 \leq x < 3) \end{cases} \qquad (9)$$

The simulation results of MATLAB and Simulink are shown in Figure below in figure 8;

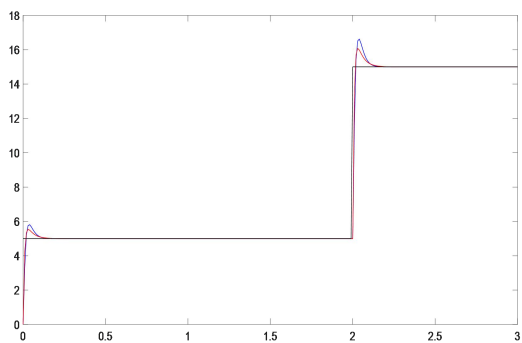

Fig. 8. Simulation data inspection of fuzzy PID controller

Table. 2 Total response data of FUZZY PID controller without interference

| PID Response Data | | | | Fuzzy PID Response Data | | | |
|---|---|---|---|---|---|---|---|
| Maximum Output Value | Peak Time (s) | Rest Time (s) | Infinity Value | Maximum Output Value | Peak Time (s) | Rest Time (s) | Infinity Value |
| 5.809 | 0.04 | 0.01889 | 5 | 5.538 | 0.03 | 0.0168 | 5 |

Here,

Peak time = The time where the output value can reach maximum, Rest time = The time when the output value can reach the infinity value.

Overshoot of the Fuzzy PID Controller is = [(Maximum output value - Infinity value)/Infinity value]×100% =[(5.538-5)/5]×100% =10.76%.

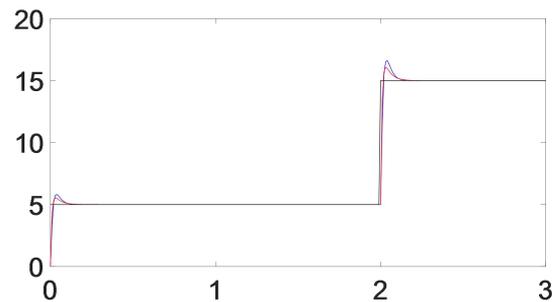

Fig. 9. Fuzzy PID output response with external interference

Table. 3 Total response data of FUZZY PID controller with external interference

| PID Response Data | | | | Fuzzy PID Response Data | | | |
|---|---|---|---|---|---|---|---|
| Maximum Output Value | Peak Time (s) | Rest Time (s) | Infinity Value | Maximum Output Value | Peak Time (s) | Rest Time (s) | Infinity Value |
| 16.618 | 2.04 | 2.018 | 15 | 16.076 | 2.03 | 2.016 | 15 |

Let's calculate the Overshoot of the Fuzzy PID Controller with external interference= [(Maximum output value - Infinity value)/Infinity value] × 100% = [(16.076-15)/15] × 100% = 7.17%

To facilitate the comparative analysis, the PID control algorithm and fuzzy PID algorithm are programmed in the PLC system. WinCC configuration is used to achieve a simple man-machine interface.

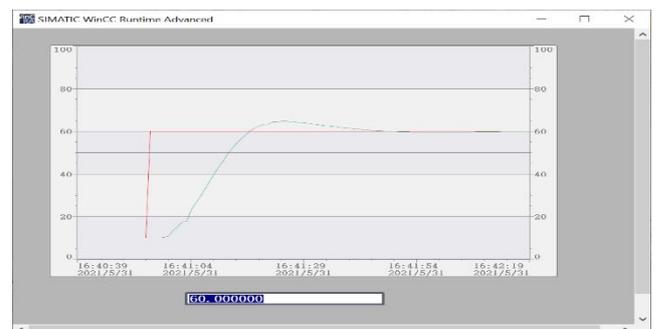

Fig. 10. PLC design output curve

Through design and research of the automatic spraying robot system and the above experimental data show that compared with the pure PID control and the fuzzy PID control has the minimum overshoot and adjusting time and rise time in the comprehensive advantages which is shown in figure 10, can better meet the requirements of automatic spraying robot control, we can reduce the manpower and material resources, make the direction of the robot according to the regulations of us ahead of time and accurate data for pesticide spraying, so fuzzy PID control system has the precision spraying output and control efficiency is very high.

## 5. Conclusion

Combined with the variable spray control system model, the simulation models of three controllers were built-in MATLAB where it can be seen from the simulation results that the overshoot of the variable spray control system controlled by the fuzzy PID controller was almost 10%, and the adjustment time was the 60s. The overshoot of the variable spraying control system is nearly 7% with external interference.

Moreover, the variable universe fuzzy PID controller also can be drawn to the smaller overshoot by adjusting the time, in the fuzzy PID controller adjusting the time is the longer, and the overshoot of the variable universe fuzzy PID controller is lesser. In the future, the next step is to design the hardware and software of the variable spray control system to further verify the variable spray control system.

**Acknowledgments**

The authors acknowledge the support from China Scholarship Council. The authors also like to thank Editor, Associate Editor, and the anonymous reviewers for constructive comments.

**Funding**

This work was partially funded by the National Key Research and Development Plan of China (No. 2018AAA0101000) and the National Natural Science Foundation of China under grant 62076028.